\documentclass[runningheads]{llncs}
\usepackage{graphicx}
\usepackage{comment}
\usepackage{amsmath,amssymb} % define this before the line numbering.
\usepackage{color}
\usepackage{booktabs}
\usepackage[width=122mm,left=12mm,paperwidth=146mm,height=193mm,top=12mm,paperheight=217mm]{geometry}
\usepackage{mytitlesec}

\begin{document}
\titlespacing*{\section} {0pt}{2pt}{2pt}
\titlespacing*{\subsection} {0pt}{2pt}{2pt}

\pagestyle{headings}
\mainmatter
\def\ECCVSubNumber{14}  % Insert your submission number here

\title{Tracking from Patterns: Learning Corresponding Patterns in Point Clouds for 3D Object Tracking} % Replace with your title

%\begin{comment}
\titlerunning{Tracking from Pattern}
\author{Jieqi Shi\inst{1}\orcidID{0000-0002-1223-191X} \and
	Peiliang Li\inst{1}\orcidID{0000-0001-5839-8777} \and
	Shaojie Shen\inst{1}\orcidID{0000-0002-5573-2909}}
\authorrunning{J. Shi et al.}
\institute{Hong Kong University of Science and Technology\\
	\email{\{jshias,pliap,eeshaojie\}@ust.hk}}	
%\end{comment}
%******************
%

\maketitle

\begin{abstract}
A robust 3D object tracker which continuously tracks surrounding objects and estimates their trajectories is key for self-driving vehicles. Most existing tracking methods employ a tracking-by-detection strategy, which usually requires complex pair-wise similarity computation and neglects the nature of continuous object motion. In this paper, we propose to directly learn 3D object correspondences from temporal point cloud data and infer the motion information from correspondence patterns. We modify the standard 3D object detector to process two lidar frames at the same time and predict bounding box pairs for the association and motion estimation tasks. We also equip our pipeline with a simple yet effective velocity smoothing module to estimate consistent object motion. Benifiting from the learned correspondences and motion refinement, our method exceeds the existing 3D tracking methods on both the KITTI and larger scale Nuscenes dataset.
\keywords{3D Detection and Tracking, Motion, Autonomous Driving}
\end{abstract}

\section{Introduction}
Traditional tracking frameworks widely adopt the tracking-by-detection strategy, where appearance features \cite{sun2019deep,frossard2018end,zhang2019robust} are utilized to calculate the correspondence map between detected objects and historic trackers. These methods treat objects as discrete items and try to find the similarity between them in an exhaustive-search manner. Though having provided some satisfying results, the traditional frameworks ignore the fact that object motion is a continunous action that happens in a local range, encoding the possible position of the object in the next moment to reduce the search range, and can be treated as identifical features. In 3D tracking, where lidar sampling is relatively random and apperance features are not stable, the ignorance of motion continuity patterns is particularly fatal.

To utilize the motion prior, some methods learn the association between frames with integrated apperance features and motion information\cite{Weng2020GNN3DMOTGN,baser2019fantrack}, but they still work on the whole frame and fail to reduce the search range with motion guidance. Others employ Kalman Filters\cite{Weng2019ABF,chiu2020probabilistic} or LSTM\cite{hu2019joint} to predict the location of objects and improve the efficiency. However, they either make a strong assumption of uniform motion or are limited to motion patterns in the training datasets, neither of which is practical in unfamiliar autonomous driving scenes where  objects move at a variable speed. To better combine the motion and apperance continuity, \cite{li2020joint,8578474,yin2020center} and \cite{zhang2019frame} concatenate two frames in the detection stage to improve both efficiency and accuracy, but they essentially stay at 2D level and can not make full use of the 3D spatial structure of point clouds. 

To solve the problem of discrete tracking and reduadant search in 3D space, we propose a 3D tracking pipeline which uses detection network to learn the correspondence patterns of two adjacent lidar frames at point level, and estimate the vehicle association and vehicle motion in the tracking module. We further reserve a sliding window for all vehicles in the subsequent tracking process to smooth the velocity. Through this pipeline, we build a universal tracking method based on correspondence patterns and solve the discontinuity in 3D tracking.
\setlength{\intextsep}{8pt} % Vertical space above & below [h] floats
\setlength{\textfloatsep}{8pt} % Vertical space below (above) [t] ([b]) floats
\begin{figure}[ht]
	\centering
	\setlength{\abovecaptionskip}{0cm} 
	\setlength{\belowcaptionskip}{-0.1cm} 
	\includegraphics[scale=0.35]{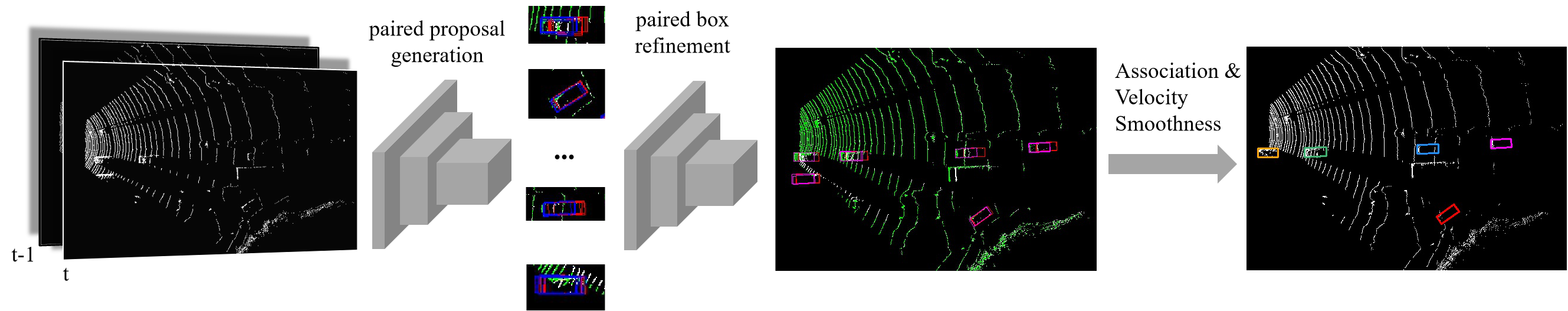}
	\caption{Illustration of our pipeline.}
	\label{fig:label}
\end{figure}
\section{Methodology and Experiments}

\subsection{Associated Detection and Tracking}
Before feeding the raw adjacent point clouds into the network, we first transform them to the same reference frame using off-the-shell estimated ego-vehicle poses\cite{qin2018vins} to decouple the camera ego-motion and learn real object motion. 

We then build our simutaneous detection and tracking module on the top of PointRCNN\cite{shi2019pointrcnn}. As illustrated in Fig.\ref{fig:label}, the detection module can be divided into two stages. In the first stage, we use the pointnet++\cite{qi2017pointnet++} to extract features for the current point cloud $pc_t$ and previous point cloud $pc_{t-1}$ respectively. A classification layer is then applied on the point-wise feature to predict an objectness score, which indicates whether this point belongs to an object in the current frame and meanwhile has correspondence in the adjacent frame. At the same time, we follow \cite{shi2019pointrcnn} to set up a regression layer and estimate two proposals for each point, with proposal consisting of seven parameters: the distance from the point to the related car, $dx, dy$ and $dz$; the orientation of the car, $ry$; and the size of the car, $h, w$ and $l$. For each point in $pc_t$ and $pc_{t-1}$, we estimate the parameters of the car in the current frame $c_t$ and the car it corresponds to in the adjacent frame $c_{t-1}$. The two stages are trained end-to-end and easy to inference.

The output of the Stage-1 network, which is in fact several 3D proposal pairs with per-point correspondence scores, form a correspondence pattern of two lidar frames in a point-to-object manner. Based on such a pattern, we build the Stage-2 network, which includes ROI Pooling and box refinement modules, following \cite{shi2019pointrcnn}, and optimize $c_t, c_{t-1}$ seperately in the Stage-2 Network to get the final paired results $c'_t$. After receiving the object pair $c'_t = \{c_t, c_{t-1}\}$, we compare the location of $c_{t-1}$ with existing trackers with simple 3D Intersection over Unions(IoUs). Noted that since we have decoupled the motion of the camera and vehicles, the pair of bounding boxes also provide us with the information of the car velocity. Considering such velocities are rough and full of bias due to detection errors, we maintain a sliding window for the velocity of cars of N most recent frames to operate first-order velocity filtering. We make the assumption that cars are moving regularly and the velocity remains steady in a certain time period, which means the sliding window helps us smooth the velocity and further helps to predict the location of cars when the detection information is lost and to fuse the new detection result with motion predictions and smooth the trajectory. 

\subsection{Experiments}
We first provide our results on the KITTI benchmark. We pretrain on the KITTI Object set without tracking images($\sim$4000 images) by moving the point clouds of each car according to a random speed. Afterwards, we split the KITTI Tracking dataset into training and validation sets following \cite{voigtlaender2019mots}. We extend the CLEAR\cite{evaluate} metrics to 3D, together with the sAMOTA defined in \cite{Weng2019ABF}, and compare our results with those of the baseline AB3DMOT\cite{Weng2019ABF} and the open-source tracking method mmMOT\cite{zhang2019robust}. For the sake of fairness, we train a PointRCNN network following the same pretaining process and use it as the detection header for both trackers. The results show that our approach outperforms each baseline. 

\begin{table}[htbp] 
	\setlength{\belowcaptionskip}{0.2cm} 
	\caption{\label{tab:test}Evaluation on KITTI Tracking \textbf{Val} Set}
	\scriptsize
	\centering
	\begin{tabular}{c|ccccccccc} 
		\toprule 
	3D IoU = 0.5  & sAMOTA $\uparrow$ & AMOTA $\uparrow$ & AMOTP$\uparrow$ & MOTA$\uparrow$ & MOTP $\uparrow$ &  IDs$\downarrow$ &  Frags$\downarrow$ &  MT$\uparrow$ & ML$\downarrow$ \\ 
		\midrule 
		AB3DMOT\cite{Weng2019ABF} & 74.45 & 30.65 & 60.22 & 63.23 & 71.75 & \bf{0.00} & \bf{141.00} & 58.67 & 8.00 \\ 
		mmMOT\cite{zhang2019robust} & 61.82 & 23.68 & 60.61 & 61.48 & \bf{71.80} & 9.00 & 283.00 & 61.33 & \bf{6.67}\\ 
		Ours & \bf{79.59} & \bf{34.37} & \bf{64.42} & \bf{65.13} & 71.29 & \bf{0.00} & 184.00 & \bf{62.00} & 7.33 \\ 
		\bottomrule
	\end{tabular} 
\end{table}

We further show the general applicability of our pipeline on the Nuscenes dataset\cite{caesar2020nuscenes}. We modify the SOTA CenterPoint\cite{yin2020center} as our detection backbone, and input two frames into the network at the same time for corresponding heatmap estimation and bounding box construction in one detector. Also, We keep the parameters in the configuration provided by CenterPoint. Our result improves from 65.0\% to 65.6\% in the overall assessment and from 81.8\% to 82.9\% on cars. 

\begin{table}[htbp] 
	\setlength{\belowcaptionskip}{0.2cm}
	\caption{\label{tab:test}Evaluation of the Tracking Pipeline on Nuscenes Test Set}
	\scriptsize
	\centering
	\begin{tabular}{c|cccccccc} 
		\toprule 
		& AMOTA $\uparrow$ & MOTA$\uparrow$ & IDs$\downarrow$ &  Frags$\downarrow$ &  MT$\uparrow$ & ML$\downarrow$ & FN $\downarrow$ & FP $\downarrow$\\ 
		\midrule 
		CenterPoint\cite{yin2020center} & 65.0 &  53.6 & \bf{684} & 553 & \bf{5399} & \bf{1818} & 24557 & 17355\\ 
		Ours & \bf{65.6} & \bf{54.3} & 732 & \bf{504} & 5383 & 1832 & \bf{24116} & \bf{16631}\\ 
		\bottomrule
		
	\end{tabular} 
\end{table}

\section{Conclusion}
In this project, we learn the point-wise correspondence pattern in the detection stage and directly give out bounding box pairs of detected cars, which solves the problem of discrete association  and reduces the search range in comparison to traditional tracking frameworks. We further provide a simple but effective method to smooth the velocity of vehicles so as to better optimize the trajectory, especially in autonomous driving tasks. Our method is simple but effective, and can be easily extened to other 3D detection models.
\clearpage
\bibliographystyle{splncs04}
\bibliography{egbib}
\end{document}